\PassOptionsToPackage{table}{xcolor}
\documentclass[11pt]{article}

\usepackage[final]{acl}
\usepackage{times}
\usepackage{latexsym}

\usepackage[T1]{fontenc}
\usepackage[utf8]{inputenc}

\usepackage{microtype}

\usepackage{inconsolata}

\usepackage[table]{xcolor}
\usepackage{graphicx} 
\usepackage{amsmath} 
\usepackage{amsfonts} 
\usepackage{booktabs} 
\usepackage{multirow} 
\usepackage{makecell} 

\usepackage{tabularx} 

\title{InstMoE: Adaptive Multimodal Routing with \\Specialized Experts}

\author{Guimin Hu$^{1}$, Xiang He$^{2}$, Yingjian Li$^{3}$,Zheng Lian$^{4}$, Boyan Xu$^{1}$, Ruichu Cai$^{1}$ \\
  $^1$Guangdong University of Technology
  $^2$University of Copenhagen \\
  $^3$Pengcheng Laboratory 
  $^4$Tongji University \\
\texttt{rice.hu.x@gmail.com}}

\begin{document}
\maketitle

\begin{abstract}
Multimodal inputs are inherently heterogeneous, not only across modalities but also in the information pathways required for effective prediction. To address this limitation, we propose \textbf{InstMoE}, an adaptive expert routing framework for multimodal learning. InstMoE dynamically routes each input to specialized unimodal and cross-modal experts, allowing the model to adapt its information pathways to the characteristics of the input. However, routing can be misled when modality-specific variations obscure task-relevant semantics. Such irrelevant variations may distort routing decisions, causing inputs to be assigned to inappropriate experts. We therefore introduce a Contrastive Semantic Alignment module, which encourages semantically similar inputs to share task-relevant representations while suppressing irrelevant modality-specific variations. Experiments on multimodal sentiment analysis benchmarks demonstrate that InstMoE achieves state-of-the-art performance on CMU-MOSEI and CH-SIMS v2 while using substantially fewer parameters than competitive baselines. Further analysis shows that different inputs exhibit distinct expert preferences, demonstrating that InstMoE moves beyond fixed fusion toward adaptive multimodal computation.
\end{abstract}

\section{Introduction}

A fundamental challenge in multimodal learning is that \textit{the importance of each modality is not fixed across inputs}. A modality that provides a reliable signal for one input may be noisy or ambiguous for another. For example, text may clearly convey a user's sentiment in some cases, whereas subtle acoustic or visual cues may provide more informative evidence when the textual signal is ambiguous. Moreover, modality-specific variations, such as speaking style, background conditions, or facial appearance, can alter the informativeness of individual modalities. This variability motivates adaptive routing that dynamically adjusts the contribution of different unimodal and cross-modal evidence pathways.

Despite this variability, most existing multimodal models rely on \textit{fixed fusion architectures}~\cite{tsai-etal-2019-multimodal, zhang-etal-2023-learning-language, HUANG2024111346}, processing different inputs through essentially the same computational pathway. This design implicitly assumes that modalities should contribute in a relatively consistent manner across samples. Such an assumption can be problematic: a dominant modality may overwhelm complementary evidence, while an unreliable modality may inject noise into the fused representation. Although prior studies have explored modality-invariant representation learning~\cite{hazarika2020misamodalityinvariantspecificrepresentations, wu2024multimodalmultilossfusionnetwork} and dynamic modality weighting~\cite{feng2024knowledgeguideddynamicmodalityattention, hu-etal-2022-unimse}, they typically focus on improving representations or adjusting modality contributions, rather than fundamentally rethinking \textit{how multimodal inputs should be processed}.

This leads to a simple but important question: \textit{should every multimodal input really follow the same fusion pathway?} We argue that an effective multimodal learner should adapt its computation to the characteristics of the input. Rather than uniformly combining all modalities, it should identify task-relevant semantic information, assess the informativeness of available modalities, and route each input to the experts that are most appropriate for its multimodal evidence. Such adaptive computation allows the model to emphasize informative modalities, suppress unreliable signals, and exploit complementary cross-modal interactions when needed.

Motivated by this perspective, we propose \textbf{InstMoE}, an \textbf{adaptive expert routing framework} that moves beyond fixed fusion through dynamic expert routing. A router dynamically determines how each input should be processed by a set of specialized modality-aware and cross-modal experts. Consequently, different inputs can activate different expert combinations and follow different multimodal processing paths, enabling the model to adapt its fusion behavior to the available evidence. InstMoE is task-agnostic and can be applied to general multimodal learning problems.

The main contributions of this work are summarized as follows:

\begin{itemize}
\item \textbf{Beyond Fixed Fusion.}
We propose \textbf{InstMoE}, an adaptive multimodal routing framework that dynamically selects among specialized unimodal and cross-modal experts according to the characteristics of each input, moving beyond the shared fusion pathways adopted by conventional multimodal models. Specifically, InstMoE identifies informative evidence pathways and adaptively combines them to construct an input-specific multimodal representation.

\item \textbf{Adaptive Evidence Routing.}
We integrate contrastive semantic alignment with dynamic expert routing, aligning multimodal representations while adaptively determining how different information sources should be processed. This design improves routing robustness by reducing the influence of irrelevant modality-specific variations.

\item \textbf{Effective and Parameter-Efficient Learning.}
InstMoE achieves state-of-the-art performance on CMU-MOSEI and CH-SIMS v2 with only 3.53M parameters, demonstrating that adaptive multimodal computation can deliver stronger performance with substantially lower model complexity.
\end{itemize}

\section{Related Work}
\label{sec:related_work}

\subsection{Multimodal Fusion}

Multimodal fusion has evolved from early and late fusion toward \textit{intermediate fusion}, which explicitly models cross-modal interactions. Existing approaches employ tensor-based fusion \cite{zadeh2017tensorfusionnetworkmultimodal, liu2018efficientlowrankmultimodalfusion}, cross-modal attention \cite{tsai-etal-2019-multimodal}, and gating or information-theoretic mechanisms \cite{rahman2020integratingmultimodalinformationlarge, han2021improvingmultimodalfusionhierarchical}. Despite their diverse designs, these methods predominantly rely on \textit{shared fusion architectures} with uniform information composition across instances. InstMoE instead enables adaptive routing over specialized modality-aware and cross-modal experts.

\subsection{Multimodal Representation Learning}

Another line of work focuses on disentangling multimodal representations to better characterize the relationships between shared and modality-specific information. MISA \cite{hazarika2020misamodalityinvariantspecificrepresentations}, for example, separates modality-invariant and modality-specific representations to preserve both shared semantics and modality-specific characteristics. Subsequent studies have explored related decomposition strategies to retain complementary modality-specific information during fusion \cite{yu2021le, wu2024multimodalmultilossfusionnetwork}. More recently, \cite{jun2025disentangling} introduces Dynamic Gaussian Anchoring to encourage attribute-separated latent units and improve the interpretability of disentangled representations, while \cite{pu2025dear} disentangles authenticity-related and event-specific representations in news content.

\subsection{Mixture of Experts (MoE)}

Mixture-of-Experts (MoE) architectures \cite{10.1162/neco.1991.3.1.79} enable conditional computation by routing inputs to specialized experts. Recent studies have extended MoE to multimodal fusion and multi-task learning \cite{wu-etal-2024-mixture, zhang2025multimodalmixturelowrankexperts}, while others have examined the specialization and interpretability of MoE routing \cite{DBLP:journals/corr/abs-2503-07639, lo2025closer}. However, existing approaches predominantly employ \textit{latent experts} whose roles are learned implicitly. InstMoE instead defines experts according to \textbf{modality-specific and cross-modal representations}, providing semantically structured choices for instance-level routing.

\section{InstMoE}
\label{chap:methodology}

\begin{table*}[t]
    \centering
    \small
    \caption{Illustrative multimodal instances showing that different instances may rely on different compositions of multimodal evidence. T, A, and V denote textual, acoustic, and visual information, respectively.}
    \label{tab:instance_examples}
    \renewcommand{\arraystretch}{1.05}
    \setlength{\tabcolsep}{5pt}
    \begin{tabularx}{\textwidth}{c|X|X|X|c}
        \toprule
        \textbf{Instance} & \textbf{Language} & \textbf{Audio} & \textbf{Vision} & \textbf{Pathway} \\
        \hline
        1 & ``I absolutely love this movie.'' & Neutral tone & Neutral expression & \textbf{T} \\
        2 & ``Yeah, that's great.'' & Sarcastic tone & Neutral expression & \textbf{T+A} \\
        3 & ``I'm fine.'' & Neutral tone & Sad expression & \textbf{T+V} \\
        4 & ``I guess I'm okay.'' & Hesitant, low-energy tone & Sad facial expression & \textbf{T+A+V} \\
        \bottomrule
    \end{tabularx}
\end{table*}

\subsection{Overview}
\label{sec:overview}

As illustrated in Table~\ref{tab:instance_examples}, different multimodal instances may require \textbf{different compositions of multimodal evidence}. While some instances can be effectively characterized by individual modalities, others require complementary information from multiple modalities and their interactions. Thus, the effective information pathway varies across instances, suggesting that a fixed fusion strategy may be suboptimal. This motivates an instance-adaptive mechanism that dynamically determines how multimodal evidence should be composed.

Building on this insight, InstMoE treats unimodal and cross-modal representations as alternative information pathways and dynamically selects the most relevant ones for each instance. Specifically, an adaptive router learns to emphasize informative pathways while suppressing less relevant ones, thereby forming an instance-specific composition of multimodal evidence. In this way, InstMoE adapts the information flow to individual instances rather than imposing a shared fusion pattern across all inputs.

\subsection{Semantic Alignment}

\paragraph{\textbf{Evidence Grounding}}
We first process the \textit{pre-extracted feature sequences} to obtain task-relevant representations for the three modalities. Specifically, the model takes aligned sequences for language ($X_t$), audio ($X_a$), and vision ($X_v$), which are derived using established pretrained feature extractors (e.g., BERT~\cite{devlin-etal-2019-bert} for language and COVAREP~\cite{COVAREP6853739} for audio). To adapt these general-purpose representations to sentiment analysis without fine-tuning the large pretrained backbones, we employ independent lightweight Transformer encoders for each modality. These encoders model intra-modal dependencies and transform the heterogeneous input features into a unified dimensionality. An attention-based aggregation layer then produces fixed-size representations $f_m$ for each modality $m\in\{t,a,v\}$.

To establish a common semantic basis for multimodal evidence, we further project the unimodal representations $f_m$ into a shared latent space using modality-specific projection heads. Let the resulting representations be denoted as $z_m$. We employ a symmetric InfoNCE objective to encourage semantically corresponding representations from different modalities to remain close while separating representations from different instances. For a text-to-modality pair $(t,m)$, where $m\in\{a,v\}$, the contrastive objective is defined as:
\begin{equation}
\mathcal{L}_{t\rightarrow m}
=
-\frac{1}{N}
\sum_{i=1}^{N}
\log
\frac{
\exp(\operatorname{sim}(z_{t,i},z_{m,i})/\tau)
}{
\sum_{j=1}^{N}
\exp(\operatorname{sim}(z_{t,i},z_{m,j})/\tau)
}
\end{equation}
where $N$ is the batch size, $(z_{t,i},z_{m,i})$ denotes the positive pair corresponding to the same instance, $\operatorname{sim}(\cdot)$ is cosine similarity, and $\tau$ is a learnable temperature parameter. We obtain the symmetric alignment objective by averaging the two directions across the auxiliary modalities:
\begin{equation}
\mathcal{L}_{\mathrm{contrastive}}
=
\frac{1}{2|\mathcal{M}|}
\sum_{m\in\mathcal{M}}
\left(
\mathcal{L}_{t\rightarrow m}
+
\mathcal{L}_{m\rightarrow t}
\right)
\end{equation}

The alignment objective is used to shape the modality encoders through backpropagation, while the resulting unimodal representations $f_t$, $f_a$, and $f_v$ are retained as direct evidence sources for subsequent evidence composition.

\subsection{Adaptive Evidence Routing}
\paragraph{\textbf{Relational Evidence}}
Unimodal representations provide direct evidence from individual modalities, but some predictive cues emerge only from interactions between modalities. We therefore construct additional relational evidence through cross-modal attention, where the textual representation queries the pre-encoded acoustic and visual feature sequences:
\begin{equation}
f_{t\rightarrow m}
=
\operatorname{Attn}(f_t,X_m),
\end{equation}
where $\operatorname{Attn}(\cdot,\cdot)$ denotes cross-modal attention with the textual representation serving as the query and the acoustic or visual features as keys and values. The resulting $f_{t\rightarrow a}$ and $f_{t\rightarrow v}$ capture interaction-aware evidence between language and acoustic or visual signals, complementing the direct evidence provided by $f_t$, $f_a$, and $f_v$. Together, these representations preserve both unimodal and relational evidence for subsequent adaptive multimodal routing.

\begin{figure}[ht!] 
    \centering
    \includegraphics[width=\linewidth]{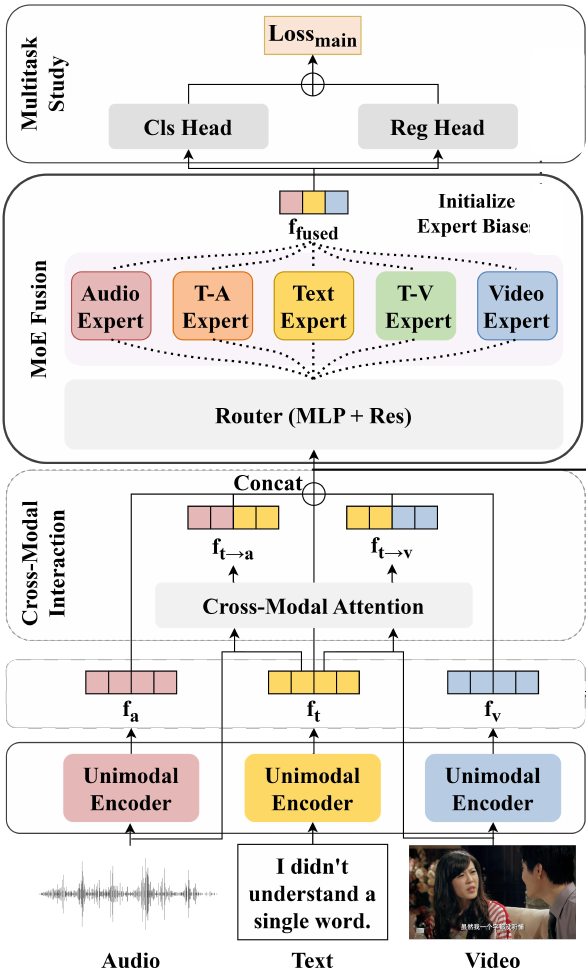}
    \caption{Overview of InstMoE.}
    \label{fig:instmoe}
\end{figure}

\paragraph{\textbf{Adaptive Multimodal Evidence Composition}}
The resulting representations define five candidate evidence pathways:
\begin{equation}
\mathcal{K}=\{t,a,v,ta,tv\},
\end{equation}
corresponding to the textual, acoustic, visual, text-audio, and text-visual evidence sources, respectively. Rather than applying a fixed fusion rule to these pathways, InstMoE assigns each pathway to a dedicated expert:
\begin{equation}
E_k(x_k)
=
\operatorname{LayerNorm}
\left(
\operatorname{ReLU}
\left(
W_kx_k+b_k
\right)
\right)
\end{equation}
where $x_k$ denotes the corresponding evidence representation and $W_k,b_k$ are trainable expert parameters.

To determine how these evidence pathways should be composed for each instance, we introduce an instance-conditioned router. The router first aggregates the available evidence:
\begin{equation}
x_{\mathrm{in}}
=
\operatorname{Concat}
\left(
f_t,f_a,f_v,f_{t\rightarrow a},f_{t\rightarrow v}
\right)
\end{equation}
and produces a probability distribution over the five evidence pathways:
\begin{equation}
\mathbf{w}
=
\operatorname{Softmax}
\left(
\mathcal{R}(x_{\mathrm{in}})
\right).
\end{equation}
where $\mathcal{R}(\cdot)$ denotes a learnable instance-conditioned routing network that maps the concatenated multimodal evidence to a set of routing logits, which are normalized by Softmax to obtain instance-specific weights over the five evidence pathways. The final multimodal representation is obtained by composing the expert outputs according to these instance-specific weights:
\begin{equation}
h_{\mathrm{moe}}
=
\sum_{k\in\mathcal{K}}
w_kE_k(x_k).
\end{equation}

The InstMoE router dynamically allocates each instance to a task-relevant combination of unimodal and cross-modal evidence pathways, moving beyond conventional dynamic modality weighting toward instance-adaptive multimodal fusion. Accordingly, the model can selectively exploit a dominant evidence source when it is sufficient for prediction, while activating complementary unimodal and cross-modal pathways when reliable inference requires richer multimodal evidence. The resulting representation $h_{\mathrm{moe}}$ is then fed into the task-specific regression and classification heads.

\paragraph{\textbf{Evidence-Aware Expert Initialization}}
To mitigate expert collapse and provide stable starting points for the evidence pathways, we initialize each expert using the empirical centroid of its corresponding training representations. Specifically, the centroid for the $k$-th evidence pathway is computed offline as:
\begin{equation}
\mathbf{c}_k
=
\frac{1}{|\mathcal{D}_k|}
\sum_{x\in\mathcal{D}_k}x,
\end{equation}
where $\mathcal{D}_k$ denotes the training representations associated with the $k$-th pathway. The resulting centroids are used to initialize the bias terms of the corresponding experts, $\mathbf{b}_k\leftarrow\mathbf{c}_k$, while the weight matrices retain standard variance-preserving initialization. The centroids are computed once before training and remain fixed throughout optimization, providing pathway-specific starting points without altering the initialization of the expert weights.

InstMoE is trained end-to-end with the final prediction objective, while auxiliary supervision is applied to the unimodal representations to preserve their task-relevant information. The contrastive alignment objective further provides semantic supervision for the common evidence space. These objectives are jointly optimized so that the model learns informative evidence pathways together with their instance-specific compositions.

\section{Experiments and Results}
\label{chap:experiments}

\subsection{Experimental Setup}
\label{sec:exp_setup}

\paragraph{Datasets} 
We evaluate the proposed InstMoE framework on two widely used multimodal benchmarks: English \textbf{CMU-MOSEI}~\cite{bagher-zadeh-etal-2018-multimodal} and Chinese \textbf{CH-SIMS v2}~\cite{liu2022makeacousticvisualcues}. Each dataset provides sequences from multiple modalities, including textual, audio, and visual information. Data splits (Train/Valid/Test) are based on official partitions. Detailed statistics and aligned/unaligned settings are provided in Table \ref{tab:dataset_stats}. We compare InstMoE against several established and recent baseline models, referenced in Section \ref{sec:main_results}. Below are brief descriptions, categorized by their core methodologies.

\begin{table}[h!]
\centering
\resizebox{\columnwidth}{!}{%
\begin{tabular}{@{} l c c l c @{}}
\toprule
\textbf{Dataset} & \textbf{Lang.} & \textbf{Total Samples} & \textbf{Split (Train/Valid/Test)} & \textbf{Alignment} \\
\midrule
CH-SIMS v2 & Zh & 4,002  & 2,400 / 800 / 802 & Unaligned \\
CMU-MOSEI  & En & 22,856 & 16,326 / 1,871 / 4,659 & Aligned \\
\bottomrule
\end{tabular}
} 
\caption{Statistics of the primary benchmark datasets.}
\label{tab:dataset_stats}
\end{table}

\paragraph{Evaluation Metrics} 
We assess performance using four standard metrics.
For regression: \textbf{Mean Absolute Error (MAE)} ($\downarrow$) and \textbf{Pearson Correlation (Corr)} ($\uparrow$).
For binary classification (positive vs. negative, excluding neutral samples): \textbf{Accuracy (Acc-2, non-zero)} ($\uparrow$) and \textbf{F1-Score (non-zero)} ($\uparrow$).

\subsection{Main Experimental Results}
\label{sec:main_results}

\subsubsection{Analysis on CH-SIMS v2 (Chinese)}
\paragraph{\textbf{Performance Analysis}}
Comparative results on the Chinese CH-SIMS v2 dataset are presented in the left of Table~\ref{tab:sims_results}. InstMoE achieves new state-of-the-art performance, outperforming all baseline models on three of the four core metrics (Corr, Acc-2, and F1-Score) and matching the state-of-the-art on MAE. Paired t-tests across five random seeds further confirm the reliability of these improvements. As indicated by the $\ddagger$ markers, InstMoE significantly outperforms the previous SOTA method, CaReFlow, with $p < 0.05$, including a 1.97-point improvement in F1-Score.

These results demonstrate the effectiveness of adaptive expert routing for multimodal sentiment analysis. Rather than applying a shared fusion strategy to all instances, InstMoE dynamically adjusts the contribution of different unimodal and cross-modal pathways according to the characteristics of each input. This behavior is consistent with our routing analysis in Sec.~\ref{sec:router_analysis}, where different sentiment conditions exhibit distinct expert preferences. Combined with its lightweight architecture, these results show that InstMoE can achieve strong multimodal performance without relying on substantially larger model capacity.

\paragraph{\textbf{Computational Efficiency Analysis}}

As summarized in the right of Table~\ref{tab:sims_results}, InstMoE achieves a highly compact design with only \textbf{3.49M} parameters, representing up to a \textbf{72\%} reduction compared to larger baselines such as Self-MMM, TETFN, and ALMT, and smaller than the recent efficient model KAN-MCP ($\sim$3.95M). 
This parameter efficiency is achieved by employing lightweight linear experts within the mixture-of-experts-style framework, in contrast to models that rely on more complex attention or spline-based computations (e.g., KAN-MCP). 
Consequently, InstMoE exhibits faster training (\textbf{65s/epoch}) and inference (\textbf{1.5ms/sample}) times than all baselines, highlighting its suitability for deployment in resource-constrained environments while maintaining superior predictive performance.

\subsubsection{Analysis on CMU-MOSEI (English)}

\paragraph{\textbf{Performance Analysis}}

Results for the large-scale English CMU-MOSEI dataset are presented in the left of Table \ref{tab:main_results_mosei}.
On this dataset, InstMoE achieves highly competitive performance, attaining a state-of-the-art F1 score of 88.67, which improves upon the previous SOTA by 1.17 points, and an Acc-2 of 87.96, also setting a new state of the art. Statistical testing further validates this result, showing a significant improvement ($p < 0.05$, marked with $\ddagger$) over the strongest baseline.
Remarkably, InstMoE achieves these superior results with significantly fewer parameters than competing Transformer-based models.
This indicates that our MoE strategy improves performance not by brute-force scaling, but by learning more efficient, specialized representations.
This suggests that with sufficient data, the MoE architecture and joint training objective facilitate the learning of robust classification capabilities.

\begin{table*}[t!]
    \centering

    \resizebox{\textwidth}{!}{%
    \begin{tabular}{l l cccc|ccc}
        \toprule
        \textbf{Model} & \textbf{Venue} & \textbf{MAE $\downarrow$} & \textbf{Corr $\uparrow$} & \textbf{Acc-2 $\uparrow$} & \textbf{F1-Score $\uparrow$}
        & \textbf{Params (M)} $\downarrow$ & \textbf{Train / ep. (s)} $\downarrow$ & \textbf{Infer. (ms)} $\downarrow$  \\
        \midrule
        
        CubeMLP*~\cite{Sun_2022}
        & ACM MM 2022
        & 0.334 & 0.648 & 71.95 & 78.53 &$\sim$6.5 & $\sim$210 & $\sim$3.8\\
        
        CENet$\dagger$~\cite{9797846}
        & TMM 2022
        & 0.310 & 0.699 & 79.56 & 79.63& $\sim$6.4 & $\sim$210 & $\sim$3.7 \\
        
        ALMT*~\cite{zhang-etal-2023-learning-language}
        & EMNLP 2023
        & 0.308 & 0.700 & 71.86 & 79.51 &$\sim$12.4&$\sim$310&$\sim$5.4\\
        KuDA*~\cite{feng2024knowledgeguideddynamicmodalityattention}
        & EMNLP 2024
        & 0.289 & 0.741 & 76.21 & 82.11& $\sim$10.6 & $\sim$280 & $\sim$5.0\\
        
        KAN-MCP*~\cite{luo2025explainablefusionbalancedlearning}
        & ACM MM 2025
        & 0.281 & 0.742 & 81.60 & 81.70 & $\sim$3.95 & $\sim$140 & $\sim$2.8\\
        
        CaReFlow~\cite{Mai_2026_CVPR}
        & CVPR 2026
        & 0.277 & 0.745 & 82.90 & 82.90 & $\sim$7 & $\sim$210 & $\sim$4.0\\
        
        CmIR~\cite{mai2026learning}
        & ACL 2026
        & 0.286 & 0.742 & 81.90 & 82.00 & $\sim$4.1 & $\sim$80 & $\sim$2.5\\
        
        \midrule
        
        \rowcolor{gray!15}
        \textbf{InstMoE}
        & \textbf{Ours}
        & \textbf{0.271}$\ddagger$
        & \textbf{0.751}$\ddagger$
        & \textbf{85.85}$\ddagger$
        & \textbf{84.87}$\ddagger$
        & \textbf{3.49}
        & \textbf{65}
        & \textbf{1.5}\\
        
        \bottomrule
    \end{tabular}
    }
    
    \caption{PA Comparative Analysis of Performance and Computational Efficiency on CH-SIMS v2. Best result per metric in \textbf{bold}. $\dagger$ denotes results from~\cite{mao2022msenaintegratedplatformmultimodal}; * denotes results from author-provided code. Results marked with $\ddagger$ indicate statistical significance ($p < 0.05$) over the previous SOTA (KAN-MCP) via paired t-test.}
    \label{tab:sims_results}
\end{table*}

\paragraph{\textbf{Computational Efficiency Analysis}}

As shown in the right of Table~\ref{tab:main_results_mosei}, InstMoE achieves superior efficiency with only \textbf{3.53M} parameters.
This represents a reduction of up to \textbf{72\%} compared to heavy baselines like MISA, and is even lighter than the recent efficient baseline KAN-MCP ($\sim$3.95M).
Crucially, InstMoE translates this compactness into speed.
Unlike architectures that rely on complex attention mechanisms or spline-based computations (e.g., KAN-MCP), our model adopts lightweight linear experts within a mixture-of-experts-style framework while achieving superior performance.
Consequently, it achieves significantly faster training (\textbf{95s/epoch}) and inference (\textbf{1.6ms/sample}) compared to all baselines, offering a robust solution for deployment-constrained environments.
\begin{table*}[t!]
\centering
\resizebox{\textwidth}{!}{%
\begin{tabular}{l l ccccc|ccc}
    \toprule
    \textbf{Model} & \textbf{Venue}
    & \textbf{MAE $\downarrow$}
    & \textbf{Corr $\uparrow$}
    & \textbf{Acc-2 $\uparrow$}
    & \textbf{F1-Score $\uparrow$}
    & \textbf{Acc-7 $\uparrow$}
    & \textbf{Params (M)} $\downarrow$
    & \textbf{Train / ep. (s)} $\downarrow$
    & \textbf{Infer. (ms)} $\downarrow$ \\
    \midrule

    UniMSE~\cite{hu-etal-2022-unimse}
    & EMNLP 2022
    & 0.523 & 0.773 & 87.46 & 87.50 & 54.39&$\sim$12.1 & $\sim$300 & $\sim$5.2 \\

    HyCon~\cite{9767560}
    & TAC 2023
    & 0.590 & 0.792 & 86.50 & 86.40 & 52.80 & $\sim$7.6 & $\sim$200 & $\sim$3.8  \\

    DMD~\cite{li2023decoupled}
    & CVPR 2023
    & - & - & 85.00&84.90& 53.70& $\sim$9.8 & $\sim$210 & $\sim$4.0\\

    MMML~\cite{wu2024multimodalmultilossfusionnetwork}
    & NAACL 2024
    & 0.517 & 0.790 & 86.49 & 86.50 & 54.95& $\sim$8.4 & $\sim$230 & $\sim$4.1 \\

    EMOE~\cite{fang2025emoe}
    & CVPR 2025
    & 0.536 & - & 85.30 & 85.30 & 54.10& $\sim$10.1 & $\sim$250 & $\sim$4.7\\

    CaReFlow~\cite{Mai_2026_CVPR}
    & CVPR 2026
    & 0.504 & \textbf{0.799} & 87.90 & 88.00 & 54.10& $\sim$7 & $\sim$210 & $\sim$4.0\\

    CmIR~\cite{mai2026learning}
    & ACL 2026
    & \textbf{0.513} & 0.793 & 87.80 & 87.70 & - &$\sim$3 & $\sim$80 & $\sim$2.5\\

    \midrule

    \rowcolor{gray!15}
    \textbf{InstMoE}
    & \textbf{Ours}
    & 0.548
    & 0.771
    & \textbf{87.96}$\ddagger$
    & \textbf{88.67}$\ddagger$
    & \textbf{55.54}
    & \textbf{3.53}
    & \textbf{95}
    & \textbf{1.6}\\

    \bottomrule
\end{tabular}
}
\caption{A Comparative Analysis of Performance and Computational Efficiency on the CMU-MOSEI Dataset. Acc-2 and F1-Score are non-zero binary metrics. Best results in \textbf{bold}. Baseline data are primarily retrieved from~\cite{wu2024multimodalmultilossfusionnetwork}, except for KAN-MCP which is cited from its original paper~\cite{luo2025explainablefusionbalancedlearning}. Results marked with $\ddagger$ indicate statistical significance ($p < 0.05$) over the second-best F1-Score.}
\label{tab:main_results_mosei}
\end{table*}

\subsection{Ablation Studies}
\label{sec:ablation_studies}


As shown in Table \ref{tab:ablation_results}, the removal of the MoE module (replaced by direct feature concatenation) caused the largest performance drop (an F1-Score drop of 3.35\%, from 84.87 to 81.52), validating the critical role of dynamic fusion.
Ablating the Semantic Alignment Loss (w/o Align Loss) led to the second-largest drop (an F1-Score drop of 2.82\%, from 84.87 to 82.05), confirming the benefit of style-content disentanglement.
The knowledge-guided initialization and auxiliary supervision losses also clearly contributed, with F1-Score drops of 1.74\% (\textbf{w/o Centroid Init}) and 1.15\% (w/o Aux Loss), respectively. The results demonstrate the effectiveness of the proposed module in multimodal sentiment analysis.

\begin{table}[ht!]
\centering
\newcommand{\cdec}[1]{{\small ($\downarrow$ #1)}} 
\newcommand{\cinc}[1]{{\small ($\uparrow$ #1)}}   

\resizebox{\columnwidth}{!}{%
\begin{tabular}{@{} l cccc @{}} 
\toprule
& \makecell{\textbf{MAE$\downarrow$}} & \makecell{\textbf{Corr.$\uparrow$}} & \makecell{\textbf{Acc-2$\uparrow$}} & \makecell{\textbf{F1-Score$\uparrow$}} \\
\midrule
\rowcolor{gray!15}
\textbf{InstMoE} & \textbf{0.271} & \textbf{0.751} & \textbf{85.85} & \textbf{84.87} \\ 
\midrule
w/o MoE Module    & 0.302 {\color{red}\cinc{0.031}} & 0.721 {\color{red}\cdec{0.030}} & 82.81 {\color{red}\cdec{3.04}} & 81.52 {\color{red}\cdec{3.35}} \\
w/o CA module    & 0.298 {\color{red}\cinc{0.027}} & 0.737 {\color{red}\cdec{0.014}} & 83.94 {\color{red}\cdec{1.91}} & 82.05 {\color{red}\cdec{2.82}} \\
w/o Centroid Init & 0.284 {\color{red}\cinc{0.013}} & 0.743 {\color{red}\cdec{0.008}} & 84.35 {\color{red}\cdec{1.50}} & 83.13 {\color{red}\cdec{1.74}} \\
w/o Aux Loss      & 0.283 {\color{red}\cinc{0.012}} & 0.747 {\color{red}\cdec{0.004}} & 84.82 {\color{red}\cdec{1.03}} & 83.72 {\color{red}\cdec{1.15}} \\
\bottomrule
\end{tabular}
} 
\caption{Ablation study results on CH-SIMS v2. The values in parentheses indicate the performance degradation compared to the full InstMoE model. ``CA'' denotes contrastive alignment module.}
\label{tab:ablation_results}
\end{table}

Collectively, these results reveal a clear hierarchy of component importance.
As demonstrated, the removal of our two core conceptual innovations, specifically dynamic fusion (w/o MoE Module) and style-content disentanglement (w/o CA), induced the most significant performance degradation.
We interpret this as strong evidence that these two components are the \textit{primary drivers} of the model's success, validating our hypotheses that models must dynamically fuse modalities and separate style from content.
The other two components (\textbf{w/o Centroid Init} and w/o Aux Loss) function as crucial \textit{supporting strategies}.
We posit they stabilize and enhance the training process, ensuring the core MoE experts specialize effectively and the unimodal features are robust before fusion.

\subsection{Router Weight Distribution Analysis}
\label{sec:router_analysis}

\begin{figure}[h!]
    \centering
    \includegraphics[width=0.9\columnwidth]{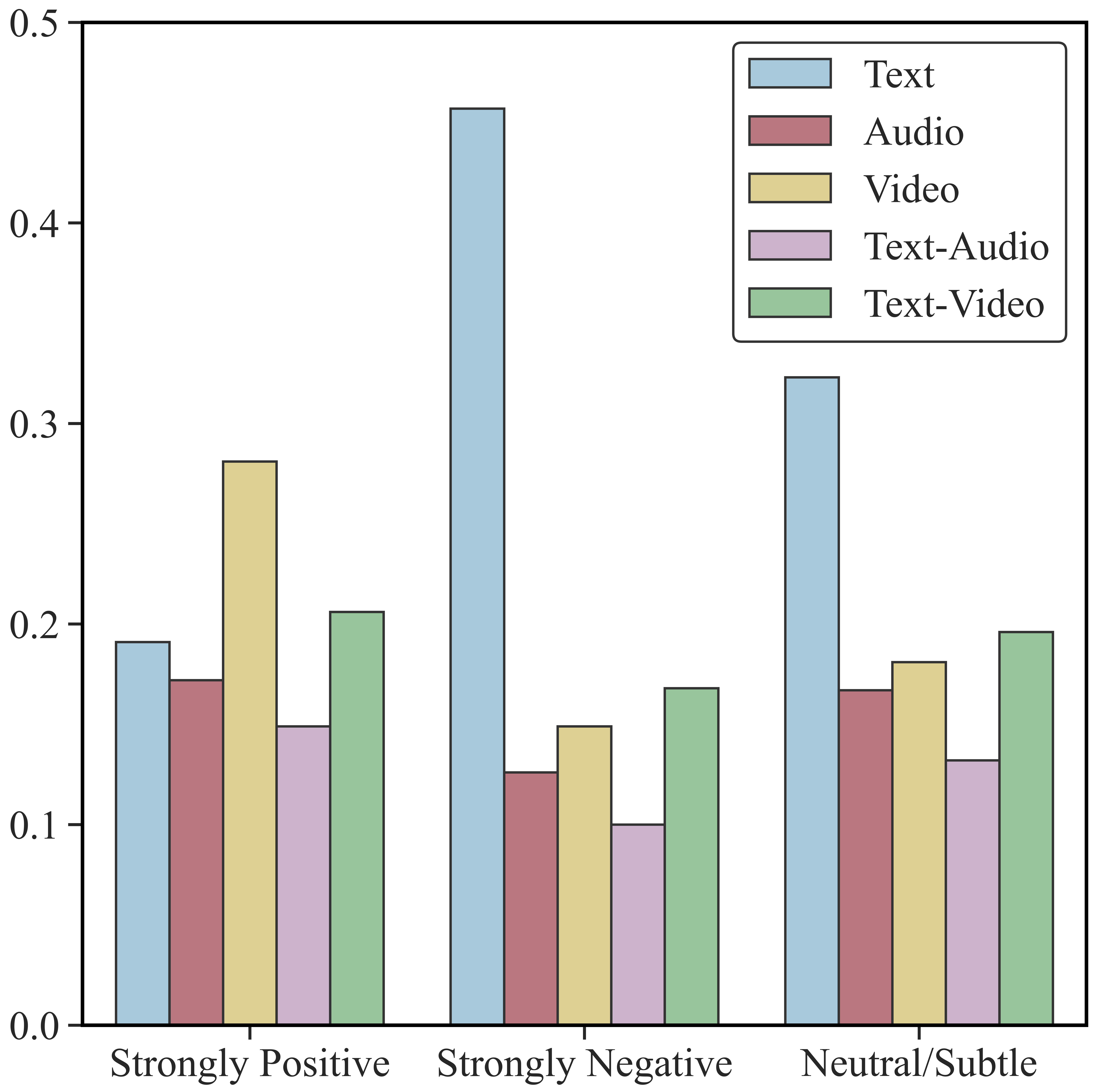}
    \caption{Average weight allocation per expert across sentiment intensity groups (CH-SIMS v2).}
    \label{fig:router_weights}
\end{figure}

To examine the router's dynamic behavior, we analyze its expert weight assignments across three sentiment intensity groups on CH-SIMS v2 (Figure~\ref{fig:router_weights}): \textit{Strongly Positive} ($\geq 0.5$), \textit{Strongly Negative} ($\leq -0.5$), and \textit{Neutral/Subtle} ($[-0.2,0.2]$).

The router exhibits clear instance-dependent variations in expert allocation. \textit{Strongly Negative} samples rely most heavily on the \textit{Text} expert (0.457), whereas \textit{Strongly Positive} samples assign greater weight to the \textit{Video} expert (0.281). In contrast, \textit{Neutral/Subtle} samples distribute substantial weight across both the \textit{Text} (0.323) and \textit{Text-Video} (0.196) experts. These shifts indicate that different sentiment conditions favor different unimodal and cross-modal pathways, rather than following a fixed fusion strategy. Overall, the results provide direct evidence that InstMoE performs instance-adaptive expert allocation, dynamically adjusting the contribution of different information pathways according to the characteristics of each input.

\section{Instance-Adaptive Routing}

\begin{table}[t]
\centering
\resizebox{\columnwidth}{!}{\begin{tabular}{lccc}
\toprule
\textbf{Condition} & $\mathbf{G_A}$ & $\mathbf{G_V}$ & $\mathbf{G_T}$ \\
\midrule
Original 
& 31.4 & 34.2 & 34.4 \\
Audio degraded (80\%) 
& 14.7 & 41.6 & 43.7 \\
Visual degraded (80\%) 
& 42.1 & 16.3 & 41.6 \\
Text degraded (80\%) 
& 43.5 & 42.7 & 13.8 \\
Audio + Visual degraded (80\%) 
& 11.2 & 13.5 & 75.3 \\
\bottomrule
\end{tabular}}
\caption{Routing behavior under controlled modality degradation. 
$G_A$, $G_V$, and $G_T$ denote the aggregated normalized routing weights 
assigned to audio-, visual-, and text-related experts, respectively. The reported values are preliminary placeholders.}
\label{tab:routing_behavior}
\end{table}

To evaluate whether InstMoE adapts its routing to modality reliability, we conduct controlled modality degradation experiments. For each instance, we selectively degrade one or more modalities while keeping the others unchanged, using temporal masking for audio and visual features and span-level masking for text at ratios of $20\%$--$80\%$. We then compare the routing weights of the original and degraded inputs using the same settings.

The routing consistently shifts away from degraded modalities. For example, degrading audio by $80\%$ reduces the weight of audio-related experts from 31.4\% to 14.7\%, while increasing the weights of visual- and text-related experts from 34.2\% and 34.4\% to 41.6\% and 43.7\%, respectively. Similar shifts occur under visual and textual degradation, with the corresponding modality weights decreasing by approximately 15--18 percentage points. Under asymmetric audio-visual degradation, the routing also responds to relative modality reliability, with the more severely degraded modality receiving a larger reduction in routing weight. These results suggest that InstMoE reallocates computation toward relatively more reliable modalities rather than merely responding to the presence of degradation.

\subsection{Representation Invariance Analysis}

To examine whether contrastive semantic alignment (CA) reduces sensitivity to irrelevant modality-specific variations while preserving task-relevant information, we train linear probes to predict speaker identity, clip source, and sentiment from representations learned with and without CA. Speaker and clip identities serve as proxies for instance-specific variations, while sentiment measures the preservation of task-relevant information. Lower speaker and clip prediction accuracy, together with stable sentiment accuracy, indicates greater representation invariance without sacrificing task-relevant information. 

As shown in Table~\ref{tab:style-probing}, CA substantially reduces the predictability of speaker and clip identity while largely preserving sentiment information. On MOSEI, for example, speaker identification accuracy decreases from 90\% to 69\%, whereas sentiment accuracy changes only slightly from 79.6\% to 78.2\%. Similar trends are observed on CH-SIMS v2. These results suggest that CA does not merely make representations invariant; it stabilizes the routing decisions against irrelevant modality-specific variation.

\begin{table}[t]
\centering
\newcommand{\cdec}[1]{{\small ($\downarrow$ #1)}}
\resizebox{\columnwidth}{!}{
\begin{tabular}{lcccc}
\toprule
\textbf{Dataset} & \textbf{Representation} & \textbf{Speaker ID} & \textbf{Clip Source} & \textbf{Emotion Acc} \\
\midrule
\multirow{2}{*}{CH-SIMS v2}
& w/o CA & 72 & 68 & 78.4 \\
&\cellcolor{gray!15} w/ CA  & \cellcolor{gray!15}\textbf{65} {\color{red}\cdec{7}} & \cellcolor{gray!15}\textbf{50} {\color{red}\cdec{18}} & \cellcolor{gray!15}\textbf{77.9} {\color{red}\cdec{0.5}} \\
\multirow{2}{*}{MOSEI}
& w/o CA & 90 & 68 & 79.6 \\
&\cellcolor{gray!15} w/ CA  & \cellcolor{gray!15}\textbf{69} {\color{red}\cdec{21}} & \cellcolor{gray!15}\textbf{54} {\color{red}\cdec{14}} & \cellcolor{gray!15}\textbf{78.2} {\color{red}\cdec{1.4}} \\
\bottomrule
\end{tabular}
}
\caption{Representation probing results with and without contrastive alignment (CA).}
\label{tab:style-probing}
\end{table}

\subsection{Perturbation Robustness}
\begin{table}[t]
\centering
\resizebox{\columnwidth}{!}{
\begin{tabular}{lccccc}
\toprule
\textbf{Dataset}&\textbf{Model} & \textbf{Pitch} $\uparrow$ & \textbf{Speed} $\uparrow$ & \textbf{Noise} $\uparrow$ & \textbf{$\Delta$Acc} $\downarrow$ \\
\midrule
\multirow{2}{*}{CH-SIMS v2}
&w/o CA & 76.4 & 74.1 & 70.2 & -5.4 \\
&\cellcolor{gray!15}w CA& \cellcolor{gray!15}\textbf{85.1} & \cellcolor{gray!15}\textbf{83.7} & \cellcolor{gray!15}\textbf{82.4} & \cellcolor{gray!15}\textbf{-2.1} \\
\multirow{2}{*}{MOSEI}
&w/o CA & 74.2 & 71.5 & 69.8 & -6.7 \\
&\cellcolor{gray!15}w/ CA & \cellcolor{gray!15}\textbf{85.4} & \cellcolor{gray!15}\textbf{86.2} & \cellcolor{gray!15}\textbf{83.8} & \cellcolor{gray!15}\textbf{-1.7} \\
\bottomrule
\end{tabular}
}
\caption{Robustness under controlled stylistic perturbations, measured by consistency between original and perturbed samples and the resulting drop in accuracy.}
\label{tab:style_perturbation}
\end{table}

To assess whether semantic alignment reduces sensitivity to irrelevant modality-specific variations, we conduct a controlled perturbation study. We apply three acoustic perturbations---pitch shifting, speed perturbation, and additive noise---to the audio modality while keeping the linguistic content unchanged. These perturbations modify acoustic characteristics without intentionally altering the underlying sentiment. For each sample, we compare the model prediction on the original and perturbed inputs. We report prediction consistency for each perturbation and the resulting accuracy change $\Delta$Acc. Higher consistency and a smaller accuracy drop indicate greater robustness to modality-specific variations.

As shown in Table~\ref{tab:style_perturbation}, models with CA consistently achieve higher prediction consistency and smaller accuracy drops across both datasets. On MOSEI, for example, consistency under pitch, speed, and noise perturbations improves from 74.2, 71.5, and 69.8 to 85.4, 86.2, and 83.8, respectively, while the accuracy drop decreases from 6.7 to 1.7 points. These results indicate that semantic alignment reduces sensitivity to superficial acoustic variations while preserving the information needed for sentiment prediction.

\begin{figure}[ht!] 
    \centering
    \includegraphics[width=0.9\columnwidth]{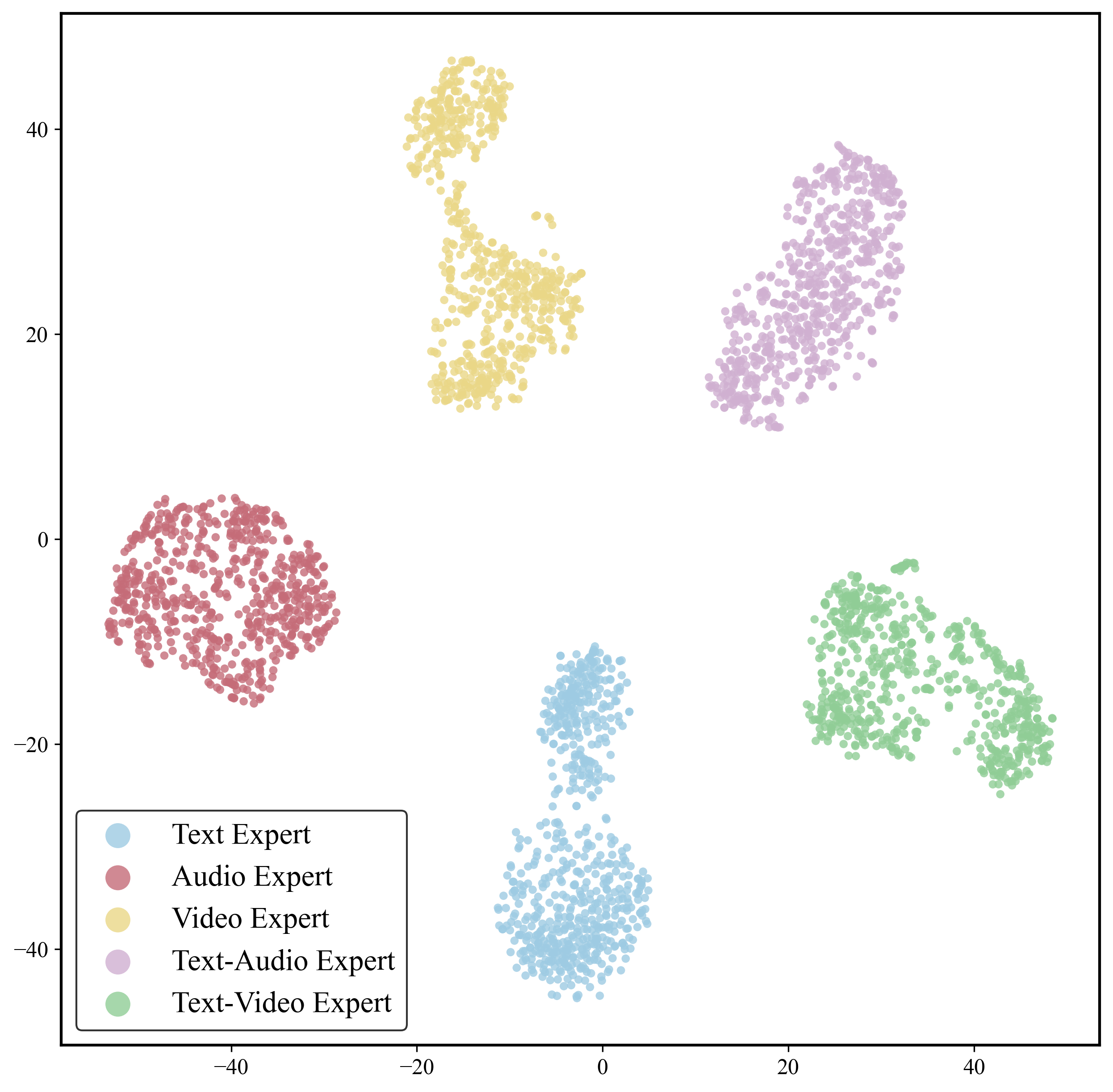} 
    \caption{t-SNE visualization of all expert network outputs (CH-SIMS v2), colored by expert type. }
    \label{fig:tsne_experts}
\end{figure}

\subsection{Expert Specialization Visualization}

To examine whether different experts learn distinct representations, we project the outputs of all five experts on CH-SIMS v2 test samples into a 2D t-SNE space, with points colored by expert identity.

As shown in Figure~\ref{fig:tsne_experts}, the five experts form clearly separated clusters with limited overlap, indicating that they learn distinct representation spaces. The separation between unimodal and cross-modal experts further suggests functional specialization, with different experts capturing modality-specific information and cross-modal interactions. These results provide qualitative evidence that the routing mechanism assigns instances to complementary rather than redundant experts, supporting the effectiveness of instance-adaptive expert specialization.

\section*{Conclusion}
We propose \textbf{InstMoE}, an instance-adaptive mixture-of-experts-style framework with three key components: (1) a \textbf{contrastive semantic alignment module} for reducing irrelevant modality-specific variations; (2) a \textbf{dynamic MoE module} with instance-adaptive routing across unimodal and cross-modal experts; and (3) a \textbf{knowledge-guided initialization strategy} for stable expert specialization. Experiments on CMU-MOSEI and CH-SIMS v2 demonstrate state-of-the-art performance with substantially fewer parameters than competitive baselines. Further analysis confirms that InstMoE learns distinct expert preferences across instances, supporting effective instance-adaptive multimodal computation.

\section*{Limitations}
Despite promising results, this study has certain limitations.
First, the current implementation relies on pre-aligned, fixed-length input sequences, potentially limiting its use in real-world scenarios with unaligned or variable-length multimodal data.
Second, our current implementation relies on standard BERT features to align with baseline protocols. 
Exploring the impact of more advanced pre-trained language models (e.g., RoBERTa, DeBERTa, or LLM-based embeddings) on expert specialization remains a promising direction for future work.
Third, style disentanglement is learned implicitly through the contrastive objective;
the lack of explicit interpretability regarding the specific features designated as "style" requires further investigation.

\section*{Ethics Statement}
We utilize publicly available datasets (CMU-MOSI,CMU-MOSEI, CH-SIMS v2) which have been scrutinized for ethical standards. However, we acknowledge that sentiment analysis models can potentially be misused for surveillance or manipulation if deployed irresponsibly. Additionally, while our model improves style disentanglement, it may still inherit biases present in the training data. We urge researchers to use this technology with discretion and primarily for analyzing aggregate user feedback or improving human-computer interaction.
\bibliography{references}

\appendix

\paragraph{\textbf{Tensor-based Fusion Models}}
\begin{itemize}
    \item TFN~\cite{zadeh2017tensorfusionnetworkmultimodal} models uni-, bi-, and tri-modal interactions via tensor (Cartesian) fusion; 
    \item LMF~\cite{liu2018efficientlowrankmultimodalfusion} improves efficiency with low-rank factorization; 
    \item 
    TETFN~\cite{10477522} further integrates text-centric tensor fusion within a Transformer framework.
\end{itemize}

\paragraph{\textbf{Transformer and Attention-based Models}}
\begin{itemize}
\item 
MulT~\cite{tsai-etal-2019-multimodal} leverages cross-modal Transformers to model interactions over unaligned sequences;
\item 
MAG-BERT~\cite{rahman2020integratingmultimodalinformationlarge} injects audio-visual cues into pretrained BERT via multimodal attention and gating; 
\item 
ALMT~\cite{zhang-etal-2023-learning-language} learns adaptive modality representations under language guidance.
\end{itemize}

\paragraph{\textbf{Disentanglement and Representation Learning}}
\begin{itemize}
    \item MISA~\cite{hazarika2020misamodalityinvariantspecificrepresentations} disentangles representations into modality-invariant and modality-specific components; 
    \item 
    Self-MM~\cite{yu2021le} learns robust modality-specific features via self-supervised multi-task learning; 
    \item MMML~\cite{wu2024multimodalmultilossfusionnetwork} extends decomposition with multiple specialized losses for optimized fusion.
\end{itemize}

\paragraph{\textbf{Other Fusion and Learning Strategies}}
\begin{itemize}
    \item 
    BBFN~\cite{han2021bi} performs pairwise bimodal fusion prior to final aggregation; 
    \item MMIM~\cite{han2021improvingmultimodalfusionhierarchical} enhances fusion via mutual information maximization; 
    \item 
    UniMSE~\cite{hu-etal-2022-unimse} provides a unified framework for MSA and MER; 
    \item 
    HyCon~\cite{9767560} leverages hybrid contrastive learning for tri-modal representation learning; 
    \item 
    CubeMLP~\cite{Sun_2022} replaces Transformers with MLP-based fusion; 
    \item 
    CENet~\cite{9797846} improves cross-modal feature enhancement; 
    \item 
    TMBL~\cite{HUANG2024111346} focuses on Transformer-based multimodal binding;
    \item KuDA~\cite{feng2024knowledgeguideddynamicmodalityattention} introduces knowledge-guided dynamic attention; \item 
    KAN-MCP~\cite{luo2025explainablefusionbalancedlearning} explores explainable fusion via KAN and multi-perspective contrastive learning.
\end{itemize}

\begin{table*}[t!]
\centering

\begin{tabularx}{0.95\textwidth}{@{} X cc @{}} 
\toprule
\textbf{Hyperparameter} & \textbf{CMU-MOSEI} & \textbf{CH-SIMS v2} \\
\midrule

\multicolumn{3}{l}{\textit{Training Parameters}} \\
Initial Learning Rate   & $6.09 \times 10^{-5}$ & $2.27 \times 10^{-4}$ \\
Weight Decay            & 0.0011                & 0.0095                \\
Max Epochs              & 50                    & 50                    \\
\midrule
\multicolumn{3}{l}{\textit{Main and Alignment Loss Weights}} \\
Alignment Loss ($\lambda_{\text{align}}$) & 0.314                 & 0.315                 \\
Main Reg. ($\lambda_{\text{reg}}$)    & 1.181                 & 2.084                 \\
Main Cls. ($\lambda_{\text{cls}}$)    & 1.960                 & 1.757                 \\
\midrule
\multicolumn{3}{l}{\textit{Auxiliary Loss Weights}} \\
Text Reg.   & 0.526                 & 0.293                 \\
Text Cls.   & 0.241                 & 0.306                 \\
Audio Reg.  & 0.053                 & 0.192                 \\
Audio Cls.  & 0.297                 & 0.022                 \\
Vision Reg. & 0.459                 & 0.029                 \\
Vision Cls. & 0.209                 & 0.010                 \\
\midrule
\multicolumn{3}{l}{\textit{Model Parameters}} \\
Modality Dropout          & 0.239                 & 0.277                 \\
Focal Loss ($\gamma$)     & 1.520                 & 2.028                 \\
Focal Loss ($\alpha_t$)   & [0.5, 0.5]            & [0.4, 0.6]            \\
\bottomrule
\end{tabularx}
\caption{Hyperparameter settings for the CMU-MOSEI and CH-SIMS v2 datasets.}
\label{tab:hyperparameters}
\end{table*}

\paragraph{Implementation Details} 
We use pre-extracted features following standard protocols in prior MSA studies~\cite{tsai-etal-2019-multimodal,hazarika2020misamodalityinvariantspecificrepresentations}. Text features are extracted with pre-trained BERT~\cite{devlin-etal-2019-bert}, while CMU-MOSEI uses COVAREP~\cite{COVAREP6853739} and Facet for audio and visual features, respectively. For CH-SIMS v2, audio and visual features are extracted using openSMILE~\cite{Eyben2010OpensmileTM} and OpenFace, respectively. Using fixed features enables us to isolate the effectiveness of the proposed \textit{InstMoE} fusion architecture.

We employ $N_{\text{exp}}=5$ experts, including three unimodal and two cross-modal experts. The router is implemented as a residual MLP with a 4-head self-attention block, followed by a two-layer MLP to produce expert weights. The model is implemented in PyTorch and trained on a single NVIDIA 4060Ti GPU using AdamW~\cite{loshchilov2019decoupledweightdecayregularization} with a batch size of 32. Dataset-specific hyperparameters, including the learning rate and semantic alignment loss weight, are tuned with Optuna~\cite{akiba2019optunanextgenerationhyperparameteroptimization}; detailed settings are reported in Table~\ref{tab:hyperparameters} in Appendix.

InstMoE uses pre-extracted features. Following standard protocols in prior MSA studies~\cite{tsai-etal-2019-multimodal,hazarika2020misamodalityinvariantspecificrepresentations}, we adopt official pre-extracted features for fair comparison. Across all datasets, text features were extracted using a pre-trained BERT model~\cite{devlin-etal-2019-bert}. Using fixed BERT representations allows us to isolate the performance gains of the proposed \textit{InstMoE} fusion architecture. For the CMU-MOSEI datasets, audio features were extracted using the COVAREP toolkit~\cite{COVAREP6853739}, while visual features were derived from Facet. For the CH-SIMS v2 dataset, audio features were extracted with the openSMILE toolkit~\cite{Eyben2010OpensmileTM}, and vision features were generated using OpenFace.

Our model architecture uses $N_{\text{exp}}=5$ experts as defined in Sec \textit{Expert Network Design} (three unimodal and two cross-modal). 
The MoE routing module is implemented as a single fusion stage following the feature encoders. The router network itself consists of a residual MLP with an attention mechanism, comprising a projection layer, a self-attention block (4 heads), and a 2-layer MLP to compute the expert weights.
Our model is implemented in PyTorch and trained on a single NVIDIA 4060Ti GPU.
We use the AdamW optimizer~\cite{loshchilov2019decoupledweightdecayregularization} with a batch size of 32. Key hyperparameters, such as the initial learning rate (e.g., $6.09e^{-5}$ for CMU-MOSEI) and the semantic alignment loss weight ($\lambda_{\text{align}}$), were optimized using Optuna~\cite{akiba2019optunanextgenerationhyperparameteroptimization} for each dataset. Key hyperparameters for CMU-MOSEI and CH-SIMS v2 are detailed in Table \ref{tab:hyperparameters}.

\end{document}